\documentclass[10pt]{article}
\usepackage[utf8]{inputenc} 
\usepackage[T1]{fontenc}    
\usepackage{hyperref}       
\usepackage{url}            
\usepackage{booktabs}       
\usepackage{amsfonts}       
\usepackage{nicefrac}       
\usepackage{microtype}      

\usepackage{latexsym}
\usepackage{graphicx}
\usepackage{amsmath}
\usepackage{amssymb}
\usepackage{amsfonts}
\usepackage{epsfig}
\usepackage{pstcol}
\usepackage{natbib,url}
\usepackage{tensor}
\usepackage{caption}
\usepackage{booktabs}
\usepackage{subcaption}
\usepackage{multirow}
\usepackage{multicol}
\usepackage{geometry}

\def\cF{{\cal F}}

\def\cL{{\cal L}}

\def\cX{{\cal X}}
\def\cY{{\cal Y}}

\def\cU{{\cal U}}

\def\cW{{\cal W}}

\def\qed{\space$\square$ \par \vspace{.15in}}

\def\hat{\widehat}

\newcommand{\br}{{\bf r}}

\newcommand{\bx}{{\bf x}}

\newcommand{\bw}{{\bf w}}

\newcommand{\bv}{{\bf v}}

\newcommand{\E}{\mbox{{\rm E}}}

\newcommand{\bc}{\begin{center}}
\newcommand{\ec}{\end{center}}
\newcommand{\be}{\begin{equation}}
\newcommand{\ee}{\end{equation}}
\newcommand{\ba}{\begin{array}}
\newcommand{\ea}{\end{array}}
\newcommand{\bean}{\begin{eqnarray*}}
\newcommand{\eean}{\end{eqnarray*}}
\newcommand{\bea}{\begin{eqnarray}}
\newcommand{\eea}{\end{eqnarray}}

\newtheorem{lemma}{\bf Lemma}

\newtheorem{proposition}{\bf Proposition}
\newtheorem{remark}{\bf Remark}

\newcommand{\ben}{\begin{enumerate}}
\newcommand{\een}{\end{enumerate}}
\newcommand{\bed}{\begin{itemize}}
\newcommand{\eed}{\end{itemize}}

\begin{document}

\title{Understanding and Improving \\Virtual Adversarial Training}

\author{%
  Dongha Kim \thanks{Department of Statistics, Seoul National University}\\
  \texttt{dongha0718@hanmail.net} \\
  Yongchan Choi \footnotemark[1]\\  
  \texttt{pminer32@gmail.com} \\
  Yongdai Kim \footnotemark[1]\\
  \texttt{ydkim0903@gmail.com} \\
}

\maketitle
\begin{abstract}
In semi-supervised learning, virtual adversarial training (\textit{VAT}) approach is one of the most attractive method due to its intuitional simplicity and powerful performances. 
\textit{VAT} finds a classifier which is robust to data perturbation toward the adversarial direction. 
In this study, we provide a fundemental explanation why \textit{VAT} works well in semi-supervised learning case and propose new techniques which are simple but powerful to improve the \textit{VAT} method.
Especially we employ the idea of \textit{Bad GAN} approach, which utilizes bad samples distributed on complement of the support of the input data, without any additional deep generative architectures.
We generate bad samples of high-quality by use of the adversarial training used in \textit{VAT} and also give theoretical explanations why the adversarial training is good at both generating bad samples.
An advantage of our proposed method is to achieve the competitive performances compared with other recent studies 
with much fewer computations.
We demonstrate advantages our method by various experiments with well known benchmark image datasets.
\end{abstract}

\section{Introduction}

Deep learning has accomplished unprecedented success due to the development of deep architectures, learning techniques and hardwares \cite{krizhevsky2012imagenet,ioffe2015batch,szegedy2015going,hinton2012improving,kingma2014adam}. However, deep learning has also suffered from collecting large amount of labeled data which requires both cost and time. Thus it becomes important to develop semi-supervised methodologies that learn a classifier (or discriminator) by using  small labeled data and large unlabeled data.

Various semi-supervised learning methods have been proposed for deep learning. 
\cite{weston2012deep} employs a manifold embedding technique using the pre-constructed graph of unlabeled data and \cite{rasmus2015semi} uses a specially designed auto-encoder to extract essential features for classification. Variational auto encoder \cite{kingma2013auto} is also used in the context of semi-supervised learning by maximizing the variational lower bound of both labeled and unlabeled data
\cite{kingma2014semi,maaloe2016auxiliary}.

Recently, a simple and powerful idea for semi-supervised learning has been proposed, which is called virtual adversarial training (\textit{VAT}, \cite{miyato2015distributional,miyato2017virtual}). 
\textit{VAT} succeeds the core idea of the adversarial training method \citep{goodfellow2014explaining} for supervised learning case, which enhances the invariance of the classifier with respect to perturbation of inputs, and apply this idea to semi-supervised learning case. 
There are several follow-up studies utilizing \textit{VAT} to sequential data or combining \textit{VAT} with their own method in order to strengthen predicting performances \citep{2016arXiv160507725M,clark2018crossview}. 
Though \textit{VAT} is intuitionally simple and has powerful performances, it is not still clear why \textit{VAT} works well in semi-supervised learning case. 

Semi-supervised learning based on generative adversarial networks (\textit{GAN}, \cite{goodfellow2014generative}) has also received much attention.
For $K$-class classification problems, \cite{salimans2016improved} solves the $(K+1)$-class classification problem where the additional $(K+1)$th class consists of synthetic images made by a generator of the \textit{GAN} learned by unlabeled data. 
\cite{dai2017good} notices that not a good generator but a bad generator which generates 
synthetic images much different from observed images is crucial, and develops a semi-supervised learning algorithm called \textit{Bad GAN}
which achieves great performances over multiple benchmark datasets.
However, \textit{Bad GAN} needs two additional deep architectures - bad generator and pre-trained density estimation model besides the one for the classifier.
 Learning these multiple deep architectures requires huge computation and memory consumption. 
In particular, the \textit{PixelCNN++} \cite{salimans2017pixelcnn++} is used for the pre-trained
density estimation model which consumes very large computational resources. 



In this study, we give fundamental explanations why \textit{VAT} works well for semi-supervised learning case.
One of the standard assumption for semi-supervised learning is \textit{cluster assumption} which means the optimal decision boundary locates at low density regions \cite{ChaSchZie06}. 
We find that \textit{VAT} pushes the decision boundary away from the high density regions of data and thus helps to find a desirable classifier given the \textit{cluster assumption}.
To be more specific, the objective function of the \textit{VAT} method can be interpreted as a differentiable version of the ideal loss function whose optimizer is a perfect classifier.


Based on the findings, we propose new techniques to enhance the performance of \textit{VAT}.
First we employ the idea of \textit{Bad GAN}, which utilizes bad samples distributed on complement of the support of input data, without any additional deep generative architectures.
Our proposed technique is motivated by close investigation of adversarial direction in \textit{VAT}. 
Here, the adversarial direction for a given datum is the direction to which the probabilities of each class change most.
We prove that the perturbed data toward their adversarial directions can serve as `good' bad samples.
\cite{dai2017good} proves that bad samples play a role to pull the decision boundary toward the low density regions of data.
By using the adversarial directions for both measuring invariance and generating the bad samples, the proposed method combines the advantages of \textit{VAT} and \textit{Bad GAN} together.
That is, our method accelerates the learning procedure by using both pushing and pulling operations simultaneously.

Secondly we modify the approximation method to calculate the adversarial direction in \cite{miyato2017virtual}.
\cite{miyato2017virtual} propose the approximation method by using the second-order Taylor expansion. 
We modify the idea by considering the reverse directions of dominant eigenvectors and find that the slight modification helps to improve the \textit{VAT} method.
We call the modified \textit{VAT} with newly proposed techniques \textit{FAT} (Fast Adversarial Training).
We show that \textit{FAT} achieves almost the state-of-the-art performances with much fewer training epochs. 
Especially, for the MNIST dataset, \textit{FAT} achieves similar test accuracies to those of \textit{Bad GAN} and  \textit{VAT}
with 5 times and 7 times fewer training epochs, respectively. 


This paper is organized as follows. In Section 2, we review the \textit{VAT} and \textit{Bad GAN} methods briefly. 
Theoretical analysis of \textit{VAT} is given in Section 3.
In Section 4, the technique to generate bad samples using the adversarial directions is described,  and our proposed semi-supervised learning method is presented.  
Results of various experiments are presented in Section 5 and conclusions follow in Section 6.

\section{Related works}

\subsection{\textit{VAT} approach}

\textit{VAT} \cite{miyato2017virtual} is a regularization method which is inspired by the adversarial training \citep{goodfellow2014explaining}. The regularization term 
of \textit{VAT} is given as:
\bean
L^{\text{VAT}}(\theta;\hat{\theta},\bx,\epsilon)&=&D_{\text{KL}}\left( p(\cdot|\bx;\hat{\theta}) || p(\cdot|\bx+\br_{\text{advr}}(\bx,\epsilon);\theta) \right)\\
&=&-\sum_{k=1}^K p(k|\bx;\hat{\theta})\log p(k|\bx+\br_{\text{advr}}(x,\epsilon);\theta)+C,
\eean
where
\be
\label{eq:advr}
\br_{\text{advr}}(\bx,\epsilon)=\underset{\br;||\br||\le\epsilon}{\text{argmax}}D_{\text{KL}}\left( p(\cdot|\bx;\hat{\theta}) || p(\cdot|\bx+\br;\hat{\theta}) \right),
\ee
$\epsilon>0$ is a tuning parameter,  $\theta$ is the parameter in the discriminator to train, $\hat{\theta}$ is the current estimate of $\theta$ and $C$ is a constant. Combining with the cross-entropy term of the labeled data, we get the final objective function of \textit{VAT}:
\bea
\label{loss:vat}
-\mathbb{E}_{\bx,y\sim\mathcal{L}^{tr}}\left[\log p(y|\bx; \theta)\right]+\mathbb{E}_{\bx\sim\mathcal{U}^{tr}}\left[L^{\text{VAT}}(\theta;\hat{\theta},\bx,\epsilon)\right],
\eea
where $\mathcal{L}^{tr}$ and $\mathcal{U}^{tr}$ are labeled and unlabeled datasets respectively.

\subsection{\textit{Bad GAN} approach}

\textit{Bad GAN} \cite{dai2017good} is a method that trains a good discriminator with a bad generator. 
Let $\mathcal{D}_G(\phi)$ be generated bad samples with a bad generator $p_G(\cdot;\phi)$ parametrized by $\phi.$ Here, the `bad generator' is a deep architecture to generate samples different from observed data. Let $p^{\text{pt}}(\cdot)$ be a pre-trained density estimation model. For a given discriminator with a feature vector $v(\bx;\theta)$
of a given input $x$ parameterized by $\theta,$
\textit{Bad GAN} learns the bad generator by minimizing the following:
$$\mathbb{E}_{\bx\sim\mathcal{D}_G(\phi)}\left[ \log p^{\text{pt}}(\bx)\mathbb{I}(p^{\text{pt}}(\bx)>\tau) \right]+|| \mathbb{E}_{\bx\sim\mathcal{U}^{tr}}v(\bx;\hat{\theta})-\mathbb{E}_{\bx\sim\mathcal{D}_G(\phi)}v(\bx;\hat{\theta})||^2$$
with respect to $\phi,$ where $\tau>0$ is a tuning parameter, $\mathcal{U}^{tr}$ is the unlabeled data, {\it and} $\hat{\theta}$ is the current estimate of $\theta$
and $\|\cdot\|$ is the Euclidean norm. 

In turn, to train the discriminator, we consider the $K$-class classification problem as the $(K+1)$-class classification problem where the $(K+1)$-th class is an artificial label of the bad samples generated by the bad generator. 
We estimate the parameter $\theta$ in the discriminator by minimizing the following:
\bea
&&-\mathbb{E}_{\bx,y\sim\mathcal{L}^{tr}}\left[\log p(y|\bx,y \leq K ; \theta)\right] - \mathbb{E}_{\bx\sim\mathcal{U}^{tr}}\left[ \log \left\{ \sum_{k=1}^K p(k|\bx;\theta)\right\} \right]\nonumber\\
&&-\mathbb{E}_{\bx\sim\mathcal{D}_G(\phi)}\left[ \log p(K+1|\bx;\theta) \right]-\mathbb{E}_{\bx\sim\mathcal{U}^{tr}}\left[\sum_{k=1}^K p(k|\bx;\theta) \log p(k|\bx;\theta) \right]
\label{eq:k1cl}
\eea
for given $\phi,$ where $\mathcal{L}^{tr}$ is the labeled set. 
See \cite{dai2017good} for details of the objective function (\ref{eq:k1cl}).

\section{Investigation of \textit{VAT} for semi-supervised learning}

In this section, we give a theoretical insight for the role of the regularization term of \textit{VAT} in semi-supervised learning. 
We show that the regularization term of \textit{VAT} pushes the decision boundary from the high density regions of unlabeled data and thus helps to find a desirable classifier given the \textit{cluster assumption}. 

\subsection{Theoretical analysis of \textit{VAT}}

Let $(\cX,d)$ be a given metric space and $(X,Y)\in\cX\times\{1,...,K\}$ be input and ouput random variables with a density function $p(\bx,y)$.
We assume that $p(\bx|y=k)$ is positive and continuous on $\cX$ and $p(y=k)>0$ for all $k$. 
For $a>0$ define $\cX(a):=\cup_{k=1}^K \cX_k(a)$ and $\delta(a):=\Pr(X\in\cX-\cX(a))$ where 
$
\cX_k(a):=\left\{\bx\in\cX: \Pr(Y=k|\bx)-\underset{k'\neq k}{\max}\Pr(Y=k'|\bx)>a \right\}.
$
And also define $P_{a}$ the probability measure of truncated random variable $X$ on $\cX(a)$ whose density function is given as $p_{a}(\bx)\propto p(\bx)\cdot \mathbb{I}(\bx\in\cX(a))$.

For a given function $f:\cX\to\mathbb{R}^K$, let $C(\cdot;f):\cX\to\{1,...,K\}$ be an induced classifier defined as $C(\bx;f):=\underset{k=1,...,K}{\text{argmax}}f_k(\bx)$ and let $C^B(\cdot)$ be the Bayes classifier. 
We denote the decision boundary of a given induced classifier $C(\cdot;f)$ by $D(f)$ and the decision boundary of the Bayes classifier by $D^B$.

Let $(X_1,Y_1),...,(X_n,Y_n)$ be $n$ independent copies from $(X,Y)$.
We propose two measures of a given function $f$: one is a standard empirical risk of labeled data given as $l_n(f):=\frac{1}{n}\sum_{i=1}^n \mathbb{I}(C(X_i;f)\neq Y_i)$,
and the other is an invariance measure of perturbation defined as
\bea
\label{inv_msr}
u_{a,\epsilon}(f):=\mathbb{E}_{P_{a}}\left[ \mathbb{I}\left( C(\bx;f)\neq C(\bx+\br;f) \text{ for all }\br\in\mathcal{B}(0,\epsilon) \right) \right],
\eea
where $a,\epsilon>0$ and $\mathcal{B}(0,\epsilon):=\{\bv:d(0,\bv)<\epsilon\}$.

Now we are ready to establish a proposition which means that the classifier which is invariant with respect to all small perturbations and classifies the labeled data correctly converges to the Bayes classifier exponentially fast with the number of labeled data. The proof is in the supplementary materials.

{\proposition{\label{vat_thm}
Let $\cF$ be a set of continuous functions including $f^B$ whose corresponding classifier is the Bayes classifier. For $a,\epsilon>0$, let $\hat{f}_n\in\underset{f\in\cF_{a,\epsilon}}{\text{argmin}}l_n(f)$, where
$
\cF_{a,\epsilon}:=\left\{ f\in\cF: f\in\underset{f\in\cF}{\text{argmin}}u_{a,\epsilon}(f) \right\}.
$
Then there exist $\delta^*,\epsilon^*,c_1^*,c_2^*>0$ not depending on $n$, if $\delta(a)<\delta^*$ and $\epsilon<\epsilon^*$, then
\bea
\label{ineq:ideal_vat}
P^{(n)}\left[ \Pr\left( C(X;\hat{f}_n)=C^B(X))\ge 1-\delta(a)  \right) \right]\ge 1-c_2^* \exp(-nc_1^*/2),
\eea
where $P^{(n)}$ is the product probability measure of $(X_1,Y_1),...,(X_n,Y_n)$.
}}

Note that the Bayes decision boundary $D^B$ locates at $\cX-\cX(a)$, hence the small value of $\delta(a)$ means essentially the \textit{cluster assumption}. 
The term $u_{a,\epsilon}(f)$ encourages the decision boundary
not to be located inside the support $\cX(a)$ 
or equivalently pushes the decision boundary from the high density regions of data, which results to find a good classifier given the \textit{cluster assumption}.

\subsection{Interpretation of \textit{VAT} }

Here we claim that the objective terms in \textit{VAT} can be interpreted as modified version of $l_n$ and $u_{\eta,\epsilon}$ respectively. 
Let $f(\bx;\theta):=(p(y=k|\bx;\theta))_{k=1}^K$ and $C(\bx;\theta):=\text{argmax}_k f_k(\bx;\theta)$. 
Proposition \ref{vat_thm} implies that it would be good to pursue a classifier which 
predicts the labeled data correctly and at the same time is invariant with respect to
all local perturbations on the unlabeled data.
 For this purpose, a plausible candidate of the objective function
is
\bea
\label{loss:cand1-re}
\mathbb{E}_{(\bx,y)\sim\mathcal{L}^{tr}}\left[ \mathbb{I}(y\neq C(\bx;\theta) \right] + \mathbb{E}_{\bx\sim\mathcal{U}^{tr}}\left[ \mathbb{I}\left(C(\bx;\theta)\neq C(\bx+\br;\theta)\text{ for all } \br\in\mathcal{B}(0,\epsilon)\right) \right],
\eea
where $\cL^{tr}$ and $\cU^{tr}$ are labeled data and unlabeled data respectively.
 

The objective function (\ref{loss:cand1-re}) is not practically usable since neither optimizing the indicator function nor checking
$C(\bx;\theta)\neq C(\bx+\br;\theta)$ for all $\br$ in $\mathcal{B}(0,\epsilon)$ is possible.
To resolve these problems, we replace the indicator functions in (\ref{loss:cand1-re}) with the cross-entropies,  
and the neighborhood $\mathcal{B}(\bx,\epsilon)$ in the second term with
the adversarial direction. By doing so, we have the following alternative objective function:
\be
\label{loss:mod1}
-\mathbb{E}_{(\bx,y)\sim\mathcal{L}^{tr}}\left[\log p(y|\bx; \theta)\right]-\mathbb{E}_{\bx\sim\mathcal{U}^{tr}}\left[\sum_{k=1}^K p(k|\bx;\theta)\log p(k|\bx+\br_{\text{advr}}(\bx,\epsilon);\theta) \right].
\ee
Finally, we replace $p(\cdot|\bx;\theta)$ in the second term of (\ref{loss:mod1}) by
 $p(\cdot|\bx;\hat{\theta})$ to have the objective function of \textit{VAT} (\ref{loss:vat}).

\section{Improved techniques for \textit{VAT}}
\label{sec:4}

\subsection{Generation of bad samples by adversarial training}

The key role of bad samples in \textit{Bad GAN} is to enforce the decision boundary to be pulled toward the low density regions of the unlabeled data. 
In this section, we propose a novel technique to generate  `good' bad samples by use of  only a given classifier.

\subsubsection{Motivation}


Let us consider the 2-class linear logistic regression model parametrized by $\eta=\{\bw,b\}$, that is, $p(y=1|x;\eta)=\left(1+\exp(-b-\bw^{'}\bx)\right)^{-1}.$
Note that the decision boundary is $\{\bx:b+\bw^{'}\bx=0\},$ and
for any given $\bx,$ the distance between $\bx$ and the decision boundary is
$|b+\bw^{'}\bx|/||\bw||.$ The key result is that moving $\bx$ toward the adversarial direction
$\br_{\text{advr}}(\bx,\epsilon)$ is equivalent to moving $\bx$ toward the decision 
boundary which is stated rigorously in the following proposition. The proof is in the supplementary materials.

{\proposition{\label{thm:2}
For a sufficiently small $\epsilon>0$, we have
\bean
\label{eq:dir}
|b+\bw^{'}\bx|/\|\bw\|>|b+\bw^{'}(\bx+ \br^{\text{advr}}(\bx,\epsilon)|/\|\bw\|
\eean
unless  $|b+\bw^{'}\bx|=0.$
}}

Hence, we can treat $\bx+C \br_{\text{advr}}(\bx,\epsilon) / ||\br^{\text{advr}}(\bx,\epsilon)||$  for appropriately choosing $C>0$
as a bad sample. 
For the decision boundary made by the DNN model with ReLU activation function, see the supplementary materials.

\subsubsection{Bad sample Generation with general classifier}\label{sec:bad}

Motivated by Proposition \ref{thm:2}, we propose
 a bad sample generator as follows. 
 Let $C>0$ be fixed and $\hat{\theta}$ be the current estimate of $\theta.$
 For a given datum $\bx$ and a classifier $p(\cdot|\bx;\hat{\theta})$, we calculate the adversarial direction  $\br_{\text{advr}}(\bx,\epsilon)$ for given $\epsilon$ by (\ref{eq:advr}). Then, we consider $\bx_{\text{bad}}=\bx+C \br_{\text{advr}}(\bx,\epsilon)/\|\br_{\text{advr}}(\bx,\epsilon)\|$ as a bad sample. 
It may happen that a generated bad sample is not sufficiently close to the decision boundary to be a 'good' bad sample, in particular when $C$ is too large or too small.
To avoid such a situation, we exclude $\bx_{\text{art}}$  which satisfies the following condition for a pre-specified $\alpha>0$:
\bean
\max_{k}p(k|\bx_{\text{bad}};\hat{\theta})>1-\alpha.
\eean

\begin{figure}[t]
\centering
\captionsetup[subfigure]{labelformat=empty}
\includegraphics[width=0.3\textwidth]{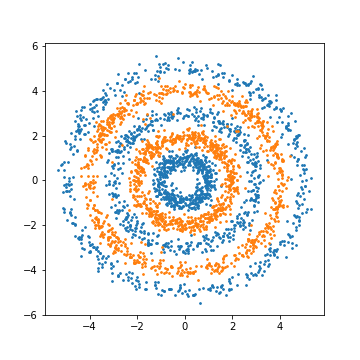}
\includegraphics[width=0.3\textwidth]{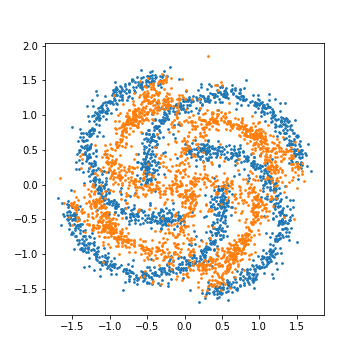}
\caption{Demonstration of how the bad samples generated by the adversarial training are distributed. We consider two cases:  3-class classification problem (\textbf{Left}) and 4-class classification problem (\textbf{Right}). 
True data and bad data are coloured by blue and orange, respectively.}
\label{fig:mul}
\end{figure}

In Figure \ref{fig:mul}, we illustrate how the bad samples generated by the proposed adversarial training are distributed for multi-class problem. With a good classifier, we can clearly see that most bad samples are located well in the low density regions of the data.

{\remark{\label{rmk:1}
Note that samples near the decision boundary are not always located in low density regions. 
Samples near the decision boundary are served as bad samples only with the reasonable classifier. 
In order to reflect this finding to learning procedure, we employ the \textit{warm-up} technique which is described in Section 4.
}}

\subsection{Modified approximation method of the adversarial direction}

\cite{miyato2017virtual} proposes the fast approximation method to calculate the adversarial direction $r_{\text{advr}}(\bx,\epsilon)$ by using the second-order Taylor expansion. 
Let us define $H(\bx,\hat{\theta})=\nabla\nabla D_{\text{KL}}\left(p(\cdot|\bx;\hat{\theta})||p(\cdot|\bx+\br;\hat{\theta})\right)|_{\br=0}$. 
They claim that $\br_{\text{advr}}$ emerges as the first dominant eigenvector $v(\bx,\hat{\theta})$ of $H(\bx,\hat{\theta})$ with magnitude $\epsilon$. 
But there always exist two dominant eigenvectors, $\pm v(\bx,\hat{\theta})$, and the sign should be selected carefully. 
So, we slightly modify the approximation method of \cite{miyato2017virtual} by 
\bean
\br_{\text{advr}}(\bx,\epsilon)=\underset{\br\in\{v(\bx,\hat{\theta}),-v(\bx,\hat{\theta})\}}{\text{argmax}}D_{\text{KL}}\left(p(\cdot|\bx;\hat{\theta})||p(\cdot|\bx+\br;\hat{\theta})\right).
\eean

\subsection{Proposed objective function}

With the new techniques described in Section \ref{sec:4}, our proposed method called \textit{FAT} updates $\theta$ by minimizing the following objective function:
\bea
&&-\mathbb{E}_{\bx,y\sim\mathcal{L}^{tr}}\left[\log p(y|\bx; \theta) \right]+
\mathbb{E}_{\bx\sim\mathcal{U}^{tr}}\left[L^{\text{VAT}}(\theta;\hat{\theta},\bx,\epsilon)\right]\nonumber\\
&&+ \lambda \cdot \left[\mathbb{E}_{\bx\sim\mathcal{U}^{tr}}\left[L^{\text{true}}(\theta;\bx)\right]+\mathbb{E}_{\bx\sim\mathcal{D}^{\text{bad}}(\hat{\theta},\epsilon,C)}\left[L^{\text{fake}}(\theta,\bx)\right] \right]
\label{loss:final}
\eea
where
$\mathcal{D}^{\text{bad}}(\hat{\theta},\epsilon,C)$ is the set of generated bad samples
with $\hat{\theta},\epsilon$ and $C,$ 
\bea
&&L^{\text{true}}(\theta;\bx)=- \sum_{k=1}^K\left[\frac{\exp(g_k(\bx;\theta))}{1+\sum_{k^{'}=1}^K \exp(g_{k^{'}}(\bx;\theta))}\log\frac{\exp(g_k(\bx;\theta))}{1+\sum_{k^{'}=1}^K \exp(g_{k^{'}}(\bx;\theta))}\right],\nonumber\\
&&L^{\text{fake}}(\theta;\bx)=-\log\frac{1}{1+\sum_{k=1}^K \exp(g_k(\bx;\theta))},\nonumber
\eea
$g(\bx;\theta)\in\mathbb{R}^K$ is a pre-softmax vector of a given architecture and $\lambda>0$.
We treat $\epsilon$ and $C$ as tuning parameters to be selected based on the validation data accuracy. 
Note that $L^{\text{fake}}$ has the same role as the sum of third and forth terms in (\ref{eq:k1cl}).

As described in Remark \ref{rmk:1}, $\bx_{\text{bad}}$s are distributed in low density regions only when the classifier performs well, which means $\bx_{\text{bad}}$s hamper the learning procedure at early learning phase.
Thus we use the \textit{warm-up} strategy \cite{bowman2015generating}, that is, we start with the small $\lambda$ being $0$ and increase it gradually after learning step proceeds.

\section{Experiments}


\subsection{Prediction performance comparisons}

We compare prediction performances of \textit{FAT} over the benchmark datasets with other semi-supervised learning algorithms. 
We consider the most widely used datasets: MNIST \citep{lecun1998gradient}, SVHN \citep{marlin2010inductive} and CIFAR10 \citep{krizhevsky2009learning}.
For fair comparison, we use the same architectures as those used in \cite{miyato2017virtual} for MNIST, SVHN and CIFAR10. 
See the supplementary materials for details.
The optimal tuning parameters $(\epsilon,C,\alpha)$ in \textit{FAT} are chosen based on the validation data accuracy.
We set $\lambda$ by zero and increase it by 0.1 after every training epoch up to one.
We use \textit{Adam} algorithm \citep{kingma2014adam} to update the parameters and do not use any data augmentation techniques. 
The results are summarized in Table \ref{tab:ts acc}, which shows that \textit{FAT} achieves the state-of-the-art accuracies for MNIST (20) and SVHN (500, 1000) and competitive accuracies with the state-of-the-art method for other settings.
Since \textit{FAT} only needs one deep architecture, we can conclude that \textit{FAT} is a powerful and computationally efficient method.
 


\begin{figure}[t]
\centering
\captionsetup[subfigure]{labelformat=empty}
\includegraphics[width=0.4\textwidth]{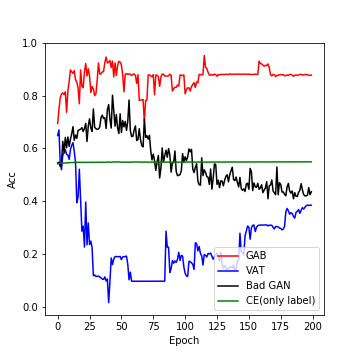}
\caption{Trace plot of the test accuracies for MNIST with 20 labeled data.}
\label{fig:mnist_small}
\end{figure}

An other advantage of \textit{FAT} is its stability with respect to learning phase. With small labeled data, Figure \ref{fig:mnist_small} shows that
the test accuracies of each epoch tends to fluctuate much and be degraded for \textit{VAT} and \textit{Bad GAN} while \textit{FAT} provides much more stable result. This may be partly because the bad samples helps to stabilize the objective function.


\begin{table*}[t]
\captionof{table}{Prediction accuracies
of various semi-supervised learning algorithms for the three benchmark datasets. $|\mathcal{L}|$ is the number of labeled data and the results with $^*$ are implemented by us. }
\begin{center}
\begin{small}
\begin{tabular}{l|ccccccc}
\hline
&\multicolumn{6}{c}{Test acc.($\%$)}\\
Data&\multicolumn{2}{c}{MNIST}&\multicolumn{2}{c}{SVHN}&\multicolumn{2}{c}{CIFAR10}\\
$|\mathcal{L}|$&20&100&500&1000&1000&4000\\
\hline
\textit{DGN} \citep{kingma2014semi}&-&96.67&-&63.98&-&-\\
\textit{Ladder} \citep{rasmus2015semi}&-&98.94&-&-&-&79.6\\
\textit{FM-GAN} \citep{salimans2016improved} &83.23&99.07&81.56&91.89&78.13&81.37\\
\textit{FM-GAN-Tan} \citep{kumar2017semi}&-&-&95.13&95.61&80.48&83.80\\
\textit{Bad GAN} \citep{dai2017good} &80.16$^*$&99.20&-&95.75&-&\bf{85.59}\\
\textit{VAT} \citep{miyato2017virtual} &67.04$^*$&98.64&-&93.17&-&85.13\\
\textit{Tri-GAN} \citep{NIPS2017_6997} &95.19&99.09&-&94.23&-&83.01\\
\textit{CCLP} \citep{pmlr-v80-kamnitsas18a} &-&\bf{99.25}&-&94.31&-&81.43\\
\hline
\textit{FAT}&\bf{96.32}&98.89&\bf{95.21}&\bf{95.94}&78.44&85.31\\
\hline
\end{tabular}
\end{small}
\label{tab:ts acc}
\end{center} 
\end{table*}

\vspace{0.5cm}
\subsection{Effects of tuning parameters}


\textit{FAT} introduces three tuning parameters $\epsilon,C$ and $\alpha$,
where $\epsilon$ is the constant used to find the adversarial direction, $C$ is the radius to generate artificial samples and $\alpha$ is used to determine whether an artificial sample is 'good'. 
We investigate the sensitivities of prediction
performances with respect to the changes of the values of these tuning parameters.  
When we vary one of the tuning parameters, the other parameters are fixed at the optimal values chosen by the validation data.
The results are reported in Table \ref{tab:tau delta}. 
Unless $\alpha$ is too small or too large, the prediction performances are not changed much.
For $\epsilon$ and $C,$ care should be done. 
With $c$ larger than the optimal $C$ (i.e. 2) or with $C$ smaller than the optimal $\epsilon$ (i.e. 1.5), the prediction performances are suboptimal.
Apparently, choosing $\epsilon$ and $C$ with $\epsilon$ being slightly smaller than $C$ gives the best result.

\begin{table}[t]
\begin{minipage}[t]{0.48\linewidth}
\captionof{table}{Test accuracies of MNIST (100) for various values of $\epsilon,C$ and $\alpha$.
The other parameters on each case are fixed to the optimal values.}
\begin{center}
\begin{small}

\begin{tabular}{cc}
\begin{tabular}{l|cccc}
\hline
$c$&1.&1.5&2.&4.\\
\hline
Test acc.&97.94&98.89&98.61&95.65\\
\hline
\end{tabular}
\\
\begin{tabular}{l|cccc}
$C$&1&2.&4.&6.\\
\hline
Test acc.&89.54&98.89&98.79&98.55\\
\hline
\end{tabular}
\\
\begin{tabular}{l|cccc}
$\alpha$&0.001&0.01&0.1&0.2\\
\hline
Test acc.&98.65&98.89&98.77&98.71\\
\hline
\end{tabular}
\end{tabular}
\end{small}

\label{tab:tau delta}
\end{center}
\end{minipage}%
    \hfill%
\begin{minipage}[t]{0.48\linewidth}
\captionof{table}{Learning time per training epoch ratios compared to supervised learning with cross-entropy for CIFAR10. \textit{Bad GAN} is operated without \textit{PixelCNN++}.}
\begin{center}

\begin{small}
\begin{tabular}{l|ccccc}
\hline
Method&\textit{VAT}&\textit{FAT}&\textit{Bad GAN}\\
\hline
Time ratio&1.37&2.09&3.20\\
\hline
\end{tabular}
\end{small}
\label{tab:learn time}
\end{center} 
\end{minipage}%
\end{table}

\vspace{0.05cm}
\subsection{Computational efficiency}

\begin{figure}[t]
\centering
\captionsetup[subfigure]{labelformat=empty}
\includegraphics[width=0.2\textwidth]{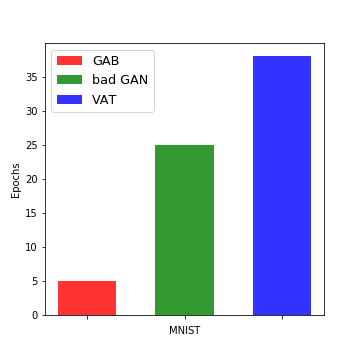}
\includegraphics[width=0.2\textwidth]{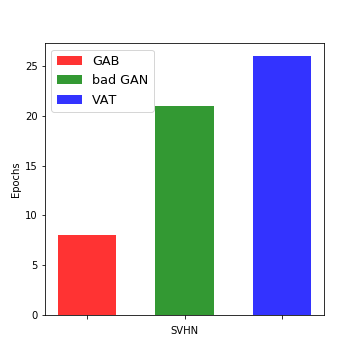}
\includegraphics[width=0.2\textwidth]{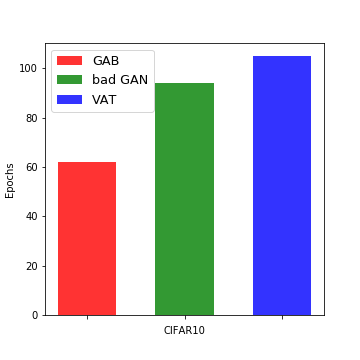}
\caption{
The number of epochs to achieve the pre-specified test accuracies (98\%, 90\% and 80\%) with the three methods for \textbf{(Left)} MNIST (100), \textbf{(Middle)} SVHN (1000) and \textbf{(Right)} CIFAR10 (4000) settings. 
\textit{Bad GAN} is operated without \textit{PixelCNN++} for SVHN and CIFAR10 datasets.}
\label{fig:learning_speed}
\end{figure}

We investigate the computational efficiency of our method in view of learning speed and computation time per training epoch. 
For \textit{Bad GAN}, we did not use \textit{PixelCNN++} on SVHN and CIFAR10 datasets since the pre-trained \textit{PixelCNN++} models are not publicly available.
Without \textit{PixelCNN++}, \textit{Bad GAN} is similar to {\it FM-GAN} \citep{salimans2016improved}.
Figure \ref{fig:learning_speed} draws the bar plots about the numbers of epochs needed to achieve the pre-specified test accuracies.
We can clearly see that \textit{FAT} requires much less epochs.

We also calculate the ratios of the computing time of each semi-supervised learning algorithm over the computing time of the corresponding supervised
learning algorithm for CIFAR10 dataset, whose results are summarized in Table \ref{tab:learn time}.
These ratios are almost same for different datasets.
The computation time of \textit{FAT} is less than \textit{Bad GAN} and competitive to \textit{VAT}.
From the results of Figure \ref{fig:learning_speed} and Table \ref{tab:learn time} we can conclude that \textit{FAT} achieves the pre-specified performances efficiently.
Note that the learning time of \textit{PixelCNN++} is not considered for this experiment,
and so comparison of computing time of \textit{FAT} and  \textit{Bad GAN} 
with \textit{PixelCNN++} is meaningless.

\subsection{Improvement of bad samples with \textit{VAT}}

For generated samples by adversarial training to be `good' bad samples,
the adversarial directions should be toward the decision boundary.
While this always happens for the linear model by Proposition \ref{thm:2}, adversarial directions could be opposite to the decision boundary for deep model.
To avoid such undesirable cases as much as possible, it would be helpful to smoothen the classifier with a regularization term. Here, we claim that the regularization term of \textit{VAT} plays such a role.

\begin{figure}[t]
\begin{minipage}[t]{0.48\linewidth}
\centering
\captionsetup[subfigure]{labelformat=empty}
\includegraphics[width=1.\textwidth]{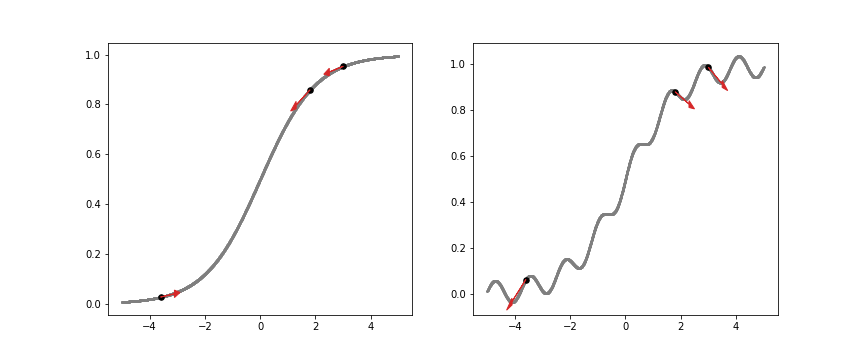} 
\caption{Examples of $P(y=1|x)$ of smooth \textbf{(Left)} and wiggle \textbf{(Right)} cases. We plot 3 points and their adversarial directions on each case.
}
\label{fig:advr_pts}
\end{minipage}%
    \hfill%
\begin{minipage}[t]{0.48\linewidth}
\centering
\captionsetup[subfigure]{labelformat=empty}
\includegraphics[width=1.\textwidth]{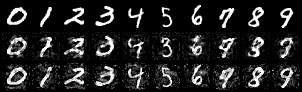} 
\caption{\textbf{(Upper)} 10 randomly sampled original MNIST dataset. \textbf{(Middle and Lower)} 
Bad samples obtained by the classifier learned with and without
the regularization term of \textit{VAT}.}
\label{fig:ce_vs_vat}
\end{minipage} 
\end{figure}

The adversarial direction obtained by maximizing the KL divergence is sensitive
to local fluctuations of the class probabilities which is examplified in
Figure \ref{fig:advr_pts}. The regularization term of \textit{VAT} is helpful to find a right adversarial direction which is toward the decision boundary by eliminating unnecessary local fluctuations of the class probabilities. 
In Figure \ref{fig:ce_vs_vat}, we compare bad samples generated by the adversarial training with and without the regularization term of \textit{VAT} for the MNIST dataset.
While the bad samples generated without the regularization term of \textit{VAT} 
are visually similar to the given input vectors, the bad samples generated with
the regularization term of \textit{VAT} look like mixtures of two different digits and thus serve as `better' bad samples.

\subsection{Quality of bad samples}
 
\begin{figure}[t]
\captionsetup[subfigure]{labelformat=empty}
\begin{center}
\hspace{.095cm}
\includegraphics[width=0.24\textwidth]{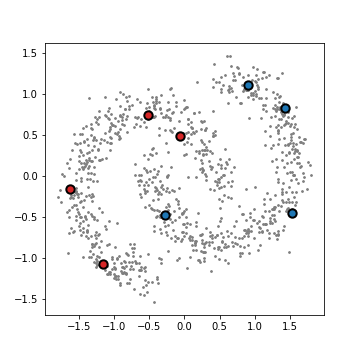} 
\includegraphics[width=0.24\textwidth]{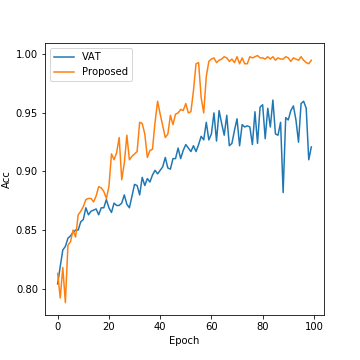}\\
\vspace{-.1cm}
\includegraphics[width=0.24\textwidth]{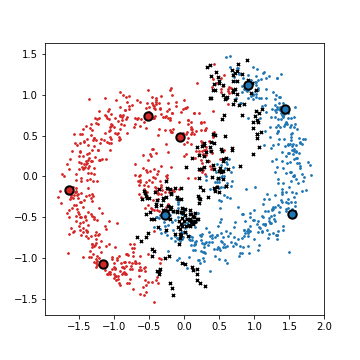}
\includegraphics[width=0.24\textwidth]{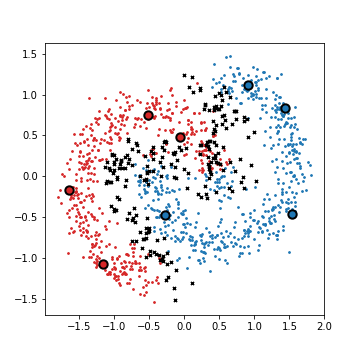}
\includegraphics[width=0.24\textwidth]{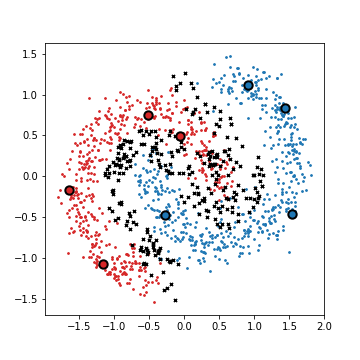}
\includegraphics[width=0.24\textwidth]{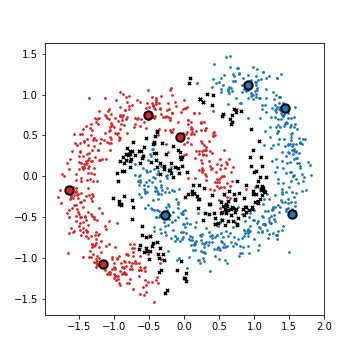}
\end{center}
\caption{\textbf{(Upper left)} The scatter plot of synthetic data which consist of 1000 unlabeled data (gray) and 4 labeled data for each class (red and blue with black edge). \textbf{(Upper right)} Accuracies of unlabeled data for each epochs for \textit{VAT} and \textit{FAT}. We use 2-layered NN with 100 hidden units each. \textbf{(Lower)} Artificial samples and classified unlabeled data by colors at the 20,40,60 and 80 training epochs of \textit{FAT} respectively.}
\label{fig:synthetic_analysis}
\end{figure}

\begin{figure}[t]
\centering
\captionsetup[subfigure]{labelformat=empty}
\includegraphics[width=0.3\textwidth]{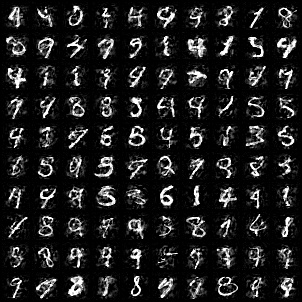}
\includegraphics[width=0.3\textwidth]{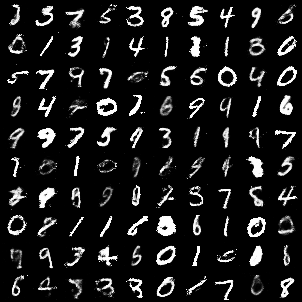}
\caption{100 randomly sampled artificial images and bad images with \textbf{(Left)} \textit{FAT} and \textbf{(Right)} \textit{Bad GAN} respectively.}
\label{fig:advr_mnist}
\end{figure}

We investigate how `good' artificial samples generated by \textit{FAT} are.
The left two plots of Figure \ref{fig:synthetic_analysis} shows the scatter plot of the synthetic data and the trace plot of prediction accuracies of \textit{FAT} and \textit{VAT}.   
And the right four plots of Figure \ref{fig:synthetic_analysis} draws the scatter plots with generated artificial samples at various epochs. 
We can clearly see that artificial samples are distributed near the current decision boundary.
We also compare artificial images generated by \textit{FAT} and bad images generated by \textit{Bad GAN} for the MNIST data at the end of learning procedure.
In Figure \ref{fig:advr_mnist},
the images by \textit{FAT} do not look like real images and do not seem to be  collapsed, which indicates that \textit{FAT} consistently generates diverse and good artificial samples. 
\textit{Bad GAN} also generates diverse bad samples but some `realistic' images can be found.

\section{Conclusion}

In this paper, we give fundamental explanations why \textit{VAT} works well for semi-supervised learning case. 
\textit{VAT} pushes the decision boundary from the high density regions of data, which results to find a good classifier given the \textit{cluster assumption}.
Also we propose a new method for semi-supervised learning, called \textit{FAT}, which modifies \textit{VAT} by introducing simple but powerful techniques.
\textit{FAT} is devised to compromise the advantages of \textit{VAT} and \textit{Bad GAN} together.
In numerical experiments, we show that \textit{FAT}
achieves almost the state-of-the-art performances with much fewer epochs.

Unlike \textit{Bad GAN} 
, \textit{FAT} only needs to learn a discriminator. Hence, it could be extended without much effort to 
other learning problems. For example, \textit{FAT} can be modified easily
for recurrent neural networks and hence can be applied to sequential data. We will
leave this extension as a future work.


\subsubsection*{Acknowledgments}
This work is supported by Samsung Electronics Co., Ltd.

\bibliographystyle{apalike}
\bibliography{reference-deep}

\section{Appendix}
\subsection{Proof of Proposition 1.}

Let $d(U,V):=\underset{u\in U,v\in V}{\min}d(u,v)$ for two sets $U$ and $V$.
For each $\cX_k(a)$, there exists a partition of open sets $\{\cX_{kj}(a)\}_{j=1}^{m_k}$ such that 
\bean
\underset{j\neq j'}{\min}d(\cX_{kj}(a),\cX_{kj'}(a))>0.
\eean
Set $\delta^*=a\cdot \min_{k,j}\Pr (X\in\cX_{kj}(a))$ and $\epsilon^*=d(\cX(a),D^B)$.
The proof consists of three steps.\\

[Step 1.]
First we show that if $f\in\cF_{a,\epsilon}$ then $d(\cX(a),D(f))\ge\epsilon$.
It is easy to check that $u_{a,\epsilon}(f^B)=0$, which means $u_{a,\epsilon}(f)=0$ for all $f\in\cF_{a,\epsilon}$. 
Let assume that $d(\cX(a),D(f))<\epsilon$
for some $f\in\cF_{a,\epsilon}$.
Then there there exists an open ball $B$ such that 
\bean
B\in \cX(a)\text{ and }d(B,D(f))<\epsilon.
\eean
Since $P_{a}(B)>0$, $u_{a,\epsilon}(f)\le 1-P_a(B)<1$, which is a contradiction. 
Therefore $d(\cX(a),D(f))\ge\epsilon$ for all $f\in\cF_{a,\epsilon}$\\

We highlight that if $f$ satisfies $d(\cX(a),D(f))\ge\epsilon$, then for any $j,k$, ${C}(\bx;f)$ is constant
on each $\mathcal{X}_{kj}(a).$\\

[Step 2.]
Let $\cW$ be the subset of $\cX^n\times\cY^n$ such that
\bea
\sum_{i=1}^n I(X_i\in\mathcal{X}_{kj}(a),Y_i=k)>\max_{k^{'}\neq k}\left\{\sum_{i=1}^n I(X_i\in\mathcal{X}_{kj}(a),Y_i=k^{'})\right\}+\sum_{i=1}^n I(X_i\in\mathcal{X}-\mathcal{X}(a))
\label{ineq:1}
\eea
for all $(k,j)$, where $\cY:=\{1,...,K\}$.
Note that $C^{B}(\bx)=k$ on $\cX_{kj}(a)$ for $j=1,...,m_k$ and $k=1,...,K$.
Hence on $\cW$,
\bean
&&\sum_{i=1}^n I(Y_i=C^{B}(X_i), X_i\in \cX_{kj}(a))
=\sum_{i=1}^n I(Y_i=k, X_i\in \cX_{kj}(a))\\
&&> \sum_{i=1}^n I(Y_i=k^{'}, X_i\in \cX_{kj}(a)) + \sum_{i=1}^n I(X_i\in\mathcal{X}-\mathcal{X}(a))
\eean
for any $k^{`}\ne k$.
 If there exists a tuple $(k,j)$ such that
$C(\bx;\hat{f})=k^{'}\ne k$ on $\bx\in \cX_{kj}(a),$ we have the following inequalities on $\cW$,
\bean
\sum_{i=1}^n (Y_i=C^{B}(X_i))\ge \sum_{i=1}^n \sum_{k,j}I(X_i\in\mathcal{X}_{k,j}(a),Y_i=k)
\eean
and
\bean
\sum_{i=1}^n I(Y_i=C(X_i;\hat{f}))&\le& \sum_{i=1}^n I(Y_i=C(X_i;\hat{f}),X_i\in\mathcal{X}(a))+\sum_{i=1}^n I(X_i\in\mathcal{X}-\mathcal{X}(a))\\
&\le& \sum_{i=1}^n\sum_{(\tilde{k},\tilde{j})\neq (k,j)}I(X_i\in\mathcal{X}_{\tilde{k}\tilde{j}}(a),Y_i=\tilde{k})+
\sum_{i=1}^n I(X_i\in\mathcal{X}_{kj}(a),Y_i=k^{'})\\
&& + \sum_{i=1}^n I(X_i\in\mathcal{X}-\mathcal{X}(a)).
\eean
Therefore, with (\ref{ineq:1}), we have
\bean
\sum_{i=1}^n (Y_i=C^{B}(X_i)) > \sum_{i=1}^n I(Y_i=C(X_i;\hat{f}))
\eean
on $\cW$, which is contradiction to the definition of $\hat{f}$. Thus
$C(\bx;\hat{f})=C^{B}(\bx)$ on $\bx\in \cX(a).$
Since $\Pr\{X\in \cX(a)\}=1-\delta(a)$
$$\Pr\left\{ C(X;\hat{f})=C^{B}(X)\right\} \ge 1-\delta(a)$$
on $\cW$.

[Step 3.]
Let $m:=\sum_{k=1}^K m_k$ and  
$$W_{i,(k,j,k^{'})}:=I(X_i\in\mathcal{X}_{kj}(a),Y_i=k)-I(X_i\in\mathcal{X}_{kj}(a),Y_i=k^{'})- I(X_i\in\mathcal{X}-\mathcal{X}(a))
$$
for $k^{'}\ne k.$ 
Note that 
\bean
\E(W_{1,(k,j,k^{'})})
&=&\int_{\mathcal{X}_{kj}(a)}\left( p(y=k|\bx)-p(y=k'|\bx)\right)p(\bx)d\bx -\delta(a)\\
&>& a \cdot \Pr(X\in\mathcal{X}_{kj}(a))-\delta(a)>0.
\eean
Let $c_1^*=Km$ and $c_2^*=\min_{k,j}\epsilon \cdot P^0(X\in\mathcal{X}_{kj}(\epsilon))-\delta(\epsilon)$. Then, Hoeffiding's inequality implies that
$$P^{(n)}\left\{\sum_{i=1}^n W_{i,(k,j,k')}< 0\right\}\le \exp(-n c_2^*/2).$$
By the union bound, we have
$$P^{(n)}(\cW^c)\le P^{(n)}\left\{\min_{k,j,k'} \sum_{i=1}^n W_{i,(k,j,k')}< 0\right\}\le
c_2^*\cdot \exp(-n c_1^*/2),$$
which complete the proof.
\qed

\subsection{Proof of Proposition 2.}

Without loss of generality, we assume that $\bw^{'}\bx+b>0$, that is, $p(y=1|\bx;\eta)>p(y=0|\bx;\eta)$. We will show that there exists $c>0$ such that $\bw^{'}\br^{\text{advr}}(\bx,\epsilon)<0$. Note that 
\bean
&&\underset{\br,||\br||\le \epsilon, \bw^{'}\br>0}{\text{argmax}}KL(\bx,\br;\eta)=\epsilon\frac{\bw}{||\bw||}(=:\br_1^{\ast})\quad\text{and}\\
&&\underset{\br,||\br||\le \epsilon, \bw^{'}\br<0}{\text{argmax}}KL(\bx,\br;\eta)=-\epsilon\frac{\bw}{||\bw||}(=:\br_2^{\ast}).
\eean
So all we have to do is to show 
$$KL(\bx,\br_2^{\ast};\eta)>KL(\bx,\br_1^{\ast};\eta).$$
By simple calculation we can get the following:
$$KL(\bx,\br_2^{\ast};\eta)-KL(\bx,\br_1^{\ast};\eta)=-\left[ p(y=1|\bx;\theta)\bw^{'}(\br_2^{\ast}-\br_1^{\ast})-\log\frac{\exp\left(\bw^{'}(\bx+\br_2^{\ast})+b\right)+1}{\exp\left(\bw^{'}(\bx+\br_1^{\ast})+b\right)+1} \right].$$
Using the Taylor's expansion up to the third-order, we obtain the following:
\bean
\log\left[\exp\left(\bw^{'}(\bx+\br)+b\right)+1\right]&=&\log\left[\exp\left(\bw^{'}\bx+b\right)+1\right]+p(y=1|\bx;\eta)\bw^{'}\br\\
&&+\frac{1}{2}p(y=1|\bx;\eta)p(y=0|\bx;\eta)(\bw^{'}\br)^2\\
&&-\frac{1}{6}p(y=1|\bx;\eta)p(y=0|\bx;\eta)\left\{p(y=1|\bx;\eta)-p(y=0|\bx;\eta)\right\}\sum_{i,j,k=1}^p w_iw_jw_k r_ir_jr_k\\
&&+o(||\br||^3).
\eean
So,
\bean
\log\frac{\exp\left(\bw^{'}(\bx+\br_2^{\ast})+b\right)+1}{\exp\left(\bw^{'}(\bx+\br_1^{\ast})+b\right)+1}&=&p(y=1|\bx;\eta)\bw^{'}(\br_2^{\ast}-\br_1^{\ast})\\
&&+\frac{1}{3}p(y=1|\bx;\eta)p(y=0|\bx;\eta)\left\{p(y=1|\bx;\eta)-p(y=0|\bx;\eta)\right\}c^3||\bw||^3+o(\epsilon^3).
\eean 
Thus, we have the following equations:
\bean
KL(\bx,\br_2^{\ast};\eta)-KL(\bx,\br_1^{\ast};\eta)&=&\frac{1}{3}p(y=1|\bx;\eta)p(y=0|\bx;\eta)\left\{p(y=1|\bx;\eta)-p(y=0|\bx;\eta)\right\}\epsilon^3||\bw||^3+o(\epsilon^3)\\
&=&C^*\cdot \epsilon^3+o(\epsilon^3).
\eean
Therefore, there exists $\epsilon^{\ast}>0$ such that $KL(\bx,\br_2^{\ast};\eta)>KL(\bx,\br_1^{\ast};\eta)$ for $\forall 0<\epsilon<\epsilon^{\ast}$. \qed

\subsection{Extension of Proposition 2 for the Deep Neural Networks}

Consider a binary classification DNN model with ReLU-like activation function
$p(y=1|\bx;\theta)=(1+\exp(-g(\bx;\theta)))^{-1}$ parameterized by $\theta.$
Here, ReLU-like function is the activation function which is piece-wise linear, such as ReLU \citep{nair2010rectified}, lReLU \citep{maas2013rectifier} and PReLU \citep{he2015delving}. 
Since $g(\bx;\theta)$ is piecewise linear,  we can write $g(\bx;\theta)$ as
\bean
g(\bx;\theta)=\sum_{j=1}^N \mathbb{I}(\bx\in\mathcal{A}_j)\cdot (\bw_j' \bx+b_j),
\eean
where $\mathcal{A}_j$ is a linear region and $N$ is the number of linear regions.

For given $\bx,$ suppose $g(\bx;\theta)>0.$ 
If $g(\bx;\theta)$ is estimated reasonably, we expect that $g(\bx;\theta)$ is decreasing if $\bx$ moves toward the decision boundary. A formal statement of this expectation would be
that $\bx - r \nabla_{\bx} g(\bx;\theta)$ can arrive at the decision boundary 
for a finite value of $r>0,$ where $\nabla_{\bx}$ is the gradient with respect to $\bx.$
 Of course, for $\bx$ with $g(\bx;\theta)<0,$
we expect that $\bx + r \nabla_{\bx} g(\bx;\theta)$ can arrive at the decision boundary 
for a finite value of $r>0.$
We say that $\bx$ is \textit{normal} if there is $r>0$ such that
$\bx-r \nabla_{\bx} g(\bx;\theta) {\rm sign} \{g(\bx;\theta)\}$ locates at the decision boundary.
We say that a linear region $\mathcal{A}_j$
is \textit{normal} if all $\bx$ in $\mathcal{A}_j$ are \textit{normal}.
We expect that most of $\mathcal{A}_j$ are \textit{normal} if $g(\bx;\theta)$ is reasonably estimated so that
the probability decreases or increases depending on ${\rm sign} \{g(\bx;\theta)\}$
if $\bx$ is getting closer to the decision boundary.

The following proposition proves that the adversarial direction is toward the decision boundary for all $\bx$s in 
\textit{normal} linear regions.

{\lemma{
If a linear region $\mathcal{A}_j$ is \textit{normal}. Then for any $\bx\in int(\mathcal{A}_j)$,
there exists $\epsilon>0$ and $C>0$ such that $\bx^A=\bx+C \br_{\text{advr}}(\bx,\epsilon)/||\br_{\text{advr}}(\bx,\epsilon)||$ is on the decision boundary.
}}
 
\textbf{Proof)} 
Take $\tilde{\epsilon}>0$ such that $\bx+\br\in\mathcal{A}_{\tilde{j}}$ for $\forall \br\in B(\bx,\tilde{\epsilon})$. 
Then by Proposition 1, there exists $0<\epsilon^{\ast}<\tilde{\epsilon}$ such that for $\forall 0<\epsilon<\epsilon^{\ast}$, 
\bean
\br_{\text{advr}}(\bx,\epsilon)&=&\epsilon\cdot \text{sign}(-b_{\tilde{j}}-\bw_{\tilde{j}}'\bx))\cdot\frac{\bw_{\tilde{j}}}{||\bw_{\tilde{j}}||}\\
&\propto&-\nabla_{\bx} g(\bx;\theta){\rm sign} \{g(\bx;\theta)\}.
\eean
$\bx$ is \textit{normal}, thus there exists $C>0$ such that $\bx^A=\bx+C \br_{\text{advr}}(\bx,\epsilon)/||\br_{\text{advr}}(\bx,\epsilon)||$ belongs to the decision boundary. \qed






\subsection{Model architectures}

All model architectures used in experiments are based on \citet{miyato2017virtual}.

\subsubsection{MNIST}
For MNIST dataset, we used fully connected NN with four hidden layers, whose numbers of nodes were (1200,600,300,150) with ReLU activation function \citep{nair2010rectified}. All the fully-connected layers are followed by BN\citep{ioffe2015batch}.

\subsubsection{SVHN, CIFAR10 and CIFAR100}
For SVHN, CIFAR10 and CIFAR10 datasets, we used the CNN architectures. More details are in Table \ref{tab:1}.

\begin{center}
\begin{tabular}{c|c|c}
\hline
SVHN&CIFAR10&CIFAR100\\
\hline
\multicolumn{3}{c}{$32\times 32$ RGB images}\\
\hline
$3\times 3$ conv. 64 lReLU&$3\times 3$ conv. 96 lReLU&$3\times 3$ conv. 128 lReLU\\
$3\times 3$ conv. 64 lReLU&$3\times 3$ conv. 96 lReLU&$3\times 3$ conv. 128 lReLU\\
$3\times 3$ conv. 64 lReLU&$3\times 3$ conv. 96 lReLU&$3\times 3$ conv. 128 lReLU\\
\hline
\multicolumn{3}{c}{$2\times 2$ max-pool, stride 2}\\
\multicolumn{3}{c}{dropout, $p=0.5$}\\
\hline
$3\times 3$ conv. 128 lReLU&$3\times 3$ conv. 192 lReLU&$3\times 3$ conv. 256 lReLU\\
$3\times 3$ conv. 128 lReLU&$3\times 3$ conv. 192 lReLU&$3\times 3$ conv. 256 lReLU\\
$3\times 3$ conv. 128 lReLU&$3\times 3$ conv. 192 lReLU&$3\times 3$ conv. 256 lReLU\\
\hline
\multicolumn{3}{c}{$2\times 2$ max-pool, stride 2}\\
\multicolumn{3}{c}{dropout, $p=0.5$}\\
\hline
$3\times 3$ conv. 128 lReLU&$3\times 3$ conv. 192 lReLU&$3\times 3$ conv. 512 lReLU\\
$1\times 1$ conv. 128 lReLU&$1\times 1$ conv. 192 lReLU&$1\times 1$ conv. 256 lReLU\\
$1\times 1$ conv. 128 lReLU&$1\times 1$ conv. 192 lReLU&$1\times 1$ conv. 128 lReLU\\
\hline
\multicolumn{3}{c}{global average pool, $6\times 6\to 1\times 1$}\\
\hline
dense $128 \to 10$&dense $192 \to 10$&dense $128 \to 100$\\
\hline
\multicolumn{2}{c}{10-way softmax}&100-way softmax\\
\hline
\end{tabular}
\captionof{table}{CNN models used in our experiments over SVHN, CIFAR10 and CIFAR100. We use leaky ReLU activation function \citep{maas2013rectifier} and all the convolutional layers and fully-connected layers are followed by BN\citep{ioffe2015batch}.}
\label{tab:1}
\end{center}
\end{document}